\documentclass{article} % For LaTeX2e
\usepackage{iclr2027_conference,times}

\usepackage{amsmath,amsfonts,bm}

\def\eqref#1{equation~\ref{#1}}
\def\1{\bm{1}}

\DeclareMathAlphabet{\mathsfit}{\encodingdefault}{\sfdefault}{m}{sl}
\SetMathAlphabet{\mathsfit}{bold}{\encodingdefault}{\sfdefault}{bx}{n}

\usepackage{url}
\usepackage{graphicx}
\usepackage{booktabs}
\usepackage{amsmath}
\usepackage{amssymb}
\usepackage{siunitx}
\usepackage{colortbl}
\usepackage[section]{placeins}
\usepackage{float}
\usepackage{needspace}
\usepackage{pgfplots}
\pgfplotsset{compat=1.18}
\usepackage{hyperref}

\definecolor{resultblue}{HTML}{3569A8}
\definecolor{resultorange}{HTML}{D17A22}
\definecolor{resultteal}{HTML}{2A8C82}
\definecolor{resultred}{HTML}{B64B4B}

\newcommand{\bench}{\textsc{IntentFlux}}
\newcommand{\gstar}{G^{\star}}
\newcommand{\gzero}{G_{0}}

\newcommand{\car}{\textsc{CAR}}
\newcommand{\idg}{\textsc{IDG}}

\newcommand{\variant}{\textsc{Variant}}
\newcommand{\decoy}{\textsc{Decoy}}
\newcommand{\stateforge}{\textsc{StateForge}}
\newcommand{\opd}{\textsc{OPD}}
\newcommand{\gentest}{\textsc{General-Test}}
\newcommand{\intertest}{\textsc{Interactive-Test}}

\title{When Users Change Their Minds: Measuring and Repairing Intent Drift in LLM Agents}

\author{Yanjie Zhang$^{1}$\thanks{Equal contribution.}\thanks{Work done during Yanjie Zhang's internship at Tencent LIGHTSPEED.}, \quad
Bowen Cao$^{2}$\footnotemark[1], \quad
Zixin Chen$^{1}$, \quad
Yushi Sun$^{3}$\thanks{Corresponding author.}\\
$^{1}$HKUST, Hong Kong SAR, China\quad\quad $^{2}$CUHK, Hong Kong SAR, China\\
$^{3}$LIGHTSPEED, Shenzhen, China\\
\texttt{\{yzhangvj,zchendf,ysunbp\}@connect.ust.hk}\quad \texttt{bwcao@link.cuhk.edu.cn}
}

\iclrfinalcopy % arXiv preprint: remove "Under review" header
\begin{document}

\maketitle
\fancyhead{} % arXiv preprint: clear the "Published as a conference paper" running head

\begin{abstract}
LLM agents often operate over multi-turn interactions in which user intent changes before execution.
We study intent drift: the failure mode in which superseded parts of the user's intent continue to influence the final answer or tool action.
We introduce \bench{}, an executable benchmark that converts verifiable tasks into dialogues with controlled intent changes while preserving their original graders.
In a 627-case calibration, mean task score falls from $0.476$ to $0.384$ as dialogues contain more superseded and withdrawn information.
Across eight models, the rate of fully correct solutions is significantly lower when the same final task must be recovered from an evolving dialogue rather than given directly in a single turn.
We further introduce \stateforge{}, which explicitly maintains the active requirements before generation.
On \gentest{}, it improves mean task score from $0.367$ to $0.467$.
Providing the ground-truth final state improves performance further but still does not recover single-turn performance, indicating that state-estimation errors explain only part of the gap.
These results establish intent drift as a measurable multi-turn failure mode and explicit state maintenance as a partial mitigation.
\end{abstract}

\section{Introduction}
\label{sec:introduction}

LLM agents often work with users who specify a task incrementally.
A user may first request an action, later revise a parameter, withdraw a
constraint, and finally ask the agent to act on the resulting
specification. The agent must then determine which parts of the user's
intent remain active. Intent drift occurs when superseded parts of the
user's intent still influence the final answer or tool action.

Existing multi-turn evaluations establish two important but incomplete
parts of this problem. Sharded-instruction settings show that models can
lose track of a task when its information is disclosed gradually
\citep{laban2025lost}. More recent evolving-intent evaluations add
argument reveal, revision, and task switching while preserving the source
verifier \citep{tack2026evolvingintent}. These settings measure whether an
agent ultimately follows an evolving task, but they do not distinguish
failures caused by obsolete intent continuing to influence the answer
from other multi-turn errors. Studying this failure directly requires a
known final intent and stale information whose erroneous use changes the
task outcome.

\bench{} addresses this gap by converting executable source tasks into
controlled multi-turn interactions in which user intent changes through
additions, deletions, and replacements. Because each interaction is
constructed from a known source task, the benchmark retains the intended
final task while controlling how earlier intent is introduced, revised,
or withdrawn. We introduce stale information in two ways: a \variant{} is
a plausible alternative value later replaced by the value in the final
intent, while a \decoy{} is a plausible constraint later withdrawn.
We retain a decoy only when obeying it changes the source-task grader's
verdict.

On a calibration pool of 627 source tasks, mean task score falls from
$0.476$ in the easy condition to $0.384$ in the hard condition, a
$19.3\%$ relative reduction. To test whether this effect is specific to
the calibration model, we evaluate eight recent LLMs and find lower
performance for the evolving-dialogue condition at every difficulty
level. A length-matched control without intent revisions scores $0.734$,
close to the $0.778$ single-turn condition and well above the
corresponding drift score of $0.469$, showing that additional turns alone
do not explain the loss. In a live tool-use environment, withholding an
explicit restatement of the final intent lowers partial-credit score by
$0.103$ relative to the clean condition, extending the effect beyond
static final-answer tasks.

We next ask whether compressing dialogue history is sufficient. We
compare two external harnesses: Deep Agents uses rolling summarization
and OpenHarness uses two-phase compaction. On \gentest{}, our fixed
125-task test set, they score $0.354$ and $0.392$, respectively, compared
with $0.367$ for bare multi-turn execution. Thus,
compression alone does not reliably recover the lost performance: the
generator may still need to determine which retained information remains
valid.

Motivated by this observation, we introduce \stateforge{}, which
separates intent-state maintenance from downstream task solving. 
A model-driven tracker removes superseded intent items and conclusions derived from them
before supplying the active state to base agent. \stateforge{} raises
mean task score on \gentest{} to $0.467$. Replacing the estimated state
with the ground-truth final intent further raises performance to $0.549$,
but still falls below the clean single-turn score of $0.778$, showing
that state-estimation errors are material but explain only part of the
remaining gap.

This decomposition motivates two complementary deployment paths:
maintaining state with a smaller trainable tracker, or internalizing the
state update into the base agent. In the modular path, a 9B tracker is
statistically indistinguishable from the 122B reference, and we use
on-policy distillation (\opd{}) to improve a 2B tracker from $0.266$ to
$0.445$. In the second path, thinking distillation internalizes
state-folding behavior into a standalone 35B agent, improving its
\gentest{} score from $0.296$ to $0.461$ without an external harness.

Our contributions are:
\begin{itemize}
  \item We introduce \bench{}, a benchmark for intent drift.
  Its controlled intent edits make stale intent measurable through its effect on source-task success.

  % \item We establish a controllable intent-drift stress condition and show that the evaluated harness baselines leave substantial errors.
  % \stateforge{} repairs part of the gap by maintaining an explicit active state before generation.
  \item We establish a controllable intent-drift stress condition and show that the evaluated history-management harnesses leave substantial errors, while \stateforge{} repairs part of the gap by maintaining an explicit active state before generation.

  \item We show that state estimation is a material component of state folding, while exact-state injection reveals residual error under the retained dialogue history.
  We further study two complementary transfer paths: distilling state maintenance into a smaller tracker and internalizing state-folding behavior into the base agent.
\end{itemize}
\section{Related Work}

\label{sec:related_work}
\vspace{-4px}
\paragraph{Evolving intent and memory updates.}
Multi-turn benchmarks study instruction retention and conversation-level
task completion \citep{kwan2024mteval,sirdeshmukh2025multichallenge,
katsis2025mtrag,laban2025lost}. EvolIF evolves per-topic constraints
through addition, deletion, and modification \citep{jia2026evolif};
InterruptBench studies revisions and retractions in long-horizon web
navigation \citep{zou2026interruptbench}; and \citet{tack2026evolvingintent}
converts verifiable tasks into reveal/revision/switch interactions.
Long-term-memory work similarly penalizes use of invalidated memories
\citep{uddin2026memora} or trains agents to prefer current over superseded
values \citep{patel2026supersede}. \bench{} complements these settings with
heterogeneous executable tasks, controlled revisions and withdrawals, and
decoys retained only when obeying them changes the source-task verdict.

\vspace{-4px}
\paragraph{Context management and state tracking.}

Long-context methods compress or retrieve growing histories
\citep{liu2023lost,jiang2023longllmlingua,lee2024readagent,
zhang2025ace}. Such methods need not explicitly resolve which parts of the
prior intent remain active. \stateforge{} instead folds the edit history
into an active requirement state before answer generation. This design also
relates to entity and belief tracking \citep{kim2023entity,
zhu2024beliefs}, but operates over open-ended task requirements and their
dependencies in executable settings.
\vspace{-4px}
\paragraph{Learning state maintenance.}

Rationale distillation and deliberative training teach intermediate
reasoning behaviors \citep{hsieh2023distilling,mukherjee2023orca,
guan2024deliberative}; on-policy distillation trains students on their
own rollouts against a teacher \citep{agarwal2023gkd}. We adapt these
ideas to two settings: the tracker sub-role and a full agent, using
executable stale-intent failures to supervise state-maintenance behavior.
\section{The \bench{} Benchmark}

\label{sec:benchmark}

\begin{figure*}[t]
\centering
\includegraphics[width=\textwidth]{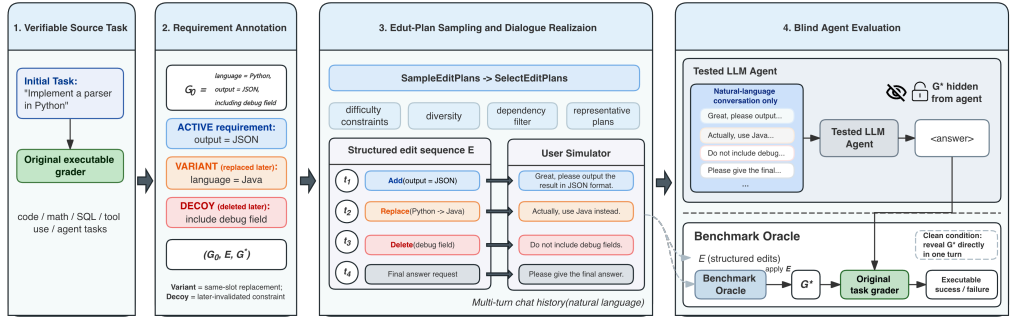}
\caption{\bench{} turns a verifiable source task into a multi-turn
interaction with controlled intent edits. The agent observes only
the dialogue; the oracle applies the edit plan to obtain the final active
intent $\gstar$, and the original grader evaluates the final output or action.}
\label{fig:intentflux-benchmark}
\end{figure*}

\bench{} measures whether an agent follows the user's \emph{current}
intent as that intent changes during a conversation. It turns
verifiable source tasks into controlled multi-turn interactions while
preserving their original graders (Figure~\ref{fig:intentflux-benchmark}).

\subsection{Evolving Intent States}

At the intent-state level, a task is represented as a finite set of
atomic natural-language goals,
$G=\{g_1,\ldots,g_n\}$. An example consists of an initial intent
state $\gzero$, a sequence of edits $E=(e_1,\ldots,e_K)$, and the final
active intent $\gstar$. The edits transform the state through
\vspace{0px}
\begin{align*}
\textsc{Add}(g) &: G_{t+1}=G_t\cup\{g\}, &
\textsc{Delete}(g) &: G_{t+1}=G_t\setminus\{g\}, \\
\textsc{Replace}(g,g') &: G_{t+1}=(G_t\setminus\{g\})\cup\{g'\}.
\end{align*}

The oracle applies the edit sequence to obtain $\gstar$, against which
the source-task grader evaluates the candidate output.

\bench{} creates stale intent in two ways. A \variant{} is a plausible
alternative value for an active intent item that is later replaced by
the target value, such as an alternative implementation language.
A \decoy{} is a plausible intent item that the user later withdraws.
We retain a decoy only when obeying it changes the source grader's
verdict. Consequently, obeying a retained decoy is grader-consequential
rather than merely a lexical mismatch. This construction makes
stale-state use testable through task success, although an aggregate
failure need not be attributable to a particular stale item without
inspecting the trajectory and output.

\subsection{Source Tasks and Construction}

The General track draws on code, math, SQL, data-to-text,
summarization, and tool-use tasks \citep{chen2021humaneval,
jain2024livecodebench,cobbe2021gsm8k,yu2018spider,parikh2020totto,
laban2024summhay,patil2025bfcl}. These tasks provide a range of output
spaces while retaining executable or task-native grading. The interactive
track uses VitaBench source tasks \citep{he2025vitabench}, where the
agent acts in delivery, in-store, OTA, and cross-domain environments.
The General track supports the stress, paired, harness, and training
experiments; VitaBench supplies the interactive-transfer tasks. Full
source-set and grader mappings appear in
Tables~\ref{tab:source-sets}--\ref{tab:graders}.

We distinguish a 627-task General-track calibration pool, a separate
502-task training pool, and two fixed evaluation sets. The calibration
pool is used only to calibrate the benchmark's difficulty response, with
the same 627 source tasks instantiated under each difficulty stratum.
The training pool is used for the OPD and thinking-distillation
experiments and is disjoint from \gentest{}. \gentest{} is a fixed,
stratified General-track test set of 125 source tasks: 10 LiveCodeBench,
21 GSM8K, 10 HumanEval, 21 Spider, 24 ToTTo, 18 SummHay, and 21 BFCL
cases. The same source cases are used at each difficulty level and for
the control and harness comparisons. \intertest{} is a fixed set of
100 VitaBench source tasks used only for interactive transfer.
Subsequent references use these set names rather than their sample counts.

At the source-task level, we decompose each problem into ordered atomic
shards and convert them into target goals that populate $\gstar$.
For most task families, source shards map directly to target goals;
task-specific loaders may instead construct goals from other source
fields as the task format requires. Variants and decoys are
additional goals introduced during intent editing and need
not correspond to source shards. For standard task families, a shard is
treated as atomic when changing it changes the task's required output
or action. We annotate behavior-changing variants and plausible decoys,
and sample edit plans subject to semantic dependency constraints.
Human annotators verify the variant, decoy, and semantic-dependency
pools; a downstream audit recovers the annotated final intent for
95\% of \gentest{}. Full annotation and quality-control procedures are
in the appendix.

\subsection{Controlled Stale-Information Load}

The General-track calibration study uses a monotone construction budget
$b\in\{1,3,5\}$ for the easy, medium, and hard strata. At budget $b$, the
sampler exposes up to $b$ variants per goal and up to $b$ eligible decoys,
under the same dependency and feasibility rules. The source cases, target
intent, source-task grader, sampling procedure, and simulator policy are
otherwise shared across strata. Because dependencies and random
initialization determine which candidates enter a realized trajectory,
the observed variant/decoy counts can be below their budgets; we report
both the schedule here and the realized means in
Table~\ref{tab:appendix-difficulty-calibration}.

Increasing $b$ jointly increases the number of superseded values and
withdrawn intent items that the model must discard. It also produces
more edit turns. We therefore treat this calibration as a controlled
test of \emph{joint stale-information load}, not as a factorial estimate
of the separate effects of variants, decoys, or length. We separately
test whether dialogue length alone can explain the degradation using a
length-matched control in Section~\ref{ssec:length-control}. After the
final edit, the simulator adds only a neutral request for the final
answer; it does not restate or summarize $\gstar$. Thus, none of the
difficulty strata receives an explicit final-intent restatement.
\section{Evaluation Protocol}

\label{sec:evaluation_protocol}

Each evaluation couples a user simulator, a tested model, and the
source task's native evaluator. The simulator realizes $(\gzero,E)$
as a multi-turn dialogue, while the final intent $\gstar$ determines
the target task scored by the source evaluator. The simulator maintains
a queue for each active goal and a separate decoy queue. At each turn,
it selects a feasible edit, expresses it as a natural user utterance,
and advances only the selected queue. This preserves the order of edits
to one goal while allowing edits to multiple goals to be naturally
interleaved.

Across all reported experiments, DeepSeek-V4-Flash serves as the user
simulator and as the rubric judge whenever an LLM-based judgment is
required. The tested agent is kept separate from these roles and varies
with the comparison.

For paired evaluations, we also construct a clean condition that reveals
$\gstar$ directly. Each task-native evaluator returns a normalized score
$s_i\in[0,1]$. We define full credit as
$c_i=\mathbf{1}[s_i=1]$ and report
\vspace{0px}
\begin{align*}
\car(\mathrm{drift}) &= \mathbb{E}[c_i\mid
\text{multi-turn drift}],\\
\car(\mathrm{clean}) &= \mathbb{E}[c_i\mid
\gstar\text{ revealed directly}],\\
\idg &= \car(\mathrm{clean})-\car(\mathrm{drift}).
\end{align*}

Thus, the current-ground-truth-intent alignment rate (\car{}) is the
fraction of the entire evaluation set receiving full task-native credit
under the current ground-truth intent, and the intent-drift gap (\idg{})
is its case-paired clean--drift difference. A positive \idg{} indicates
lower task completion when the same final task must be recovered from
an evolving interaction rather than presented directly. Clean and drift
conditions share the source case, final intent, grader, and tested model,
holding task identity and model capability fixed across the pair. The
conditions intentionally differ in interaction form, however; \idg{}
captures the total performance loss associated with recovering the final
intent from the dialogue rather than isolating a single causal factor.
We therefore interpret \idg{} together with a separate turn-matched
no-drift control, whose narrower purpose is to test whether additional
turns alone account for the loss. Key comparisons use case-paired
bootstrap confidence intervals.

Code, math, database, and tool-use scores are binary. ToTTo and SummHay
produce continuous task-native scores; for \car{}, they count as success
only at full credit ($s_i=1$), with no tuned threshold. They remain in
the denominator, so \car{} on \gentest{} always uses all 125 expected
cases, not only the 83 binary-task cases. \car{} therefore serves as a
strict full-task-completion measure, while mean score captures partial
credit on continuous-output tasks. If the tested model fails to produce
a scorable output, the case receives $s_i=c_i=0$ and remains in the
denominator.

We separately report the case-micro mean
$N^{-1}\sum_i s_i$, denoted \emph{mean score}. Because this average mixes
task-native metrics, we use it only for within-benchmark comparisons under
the fixed task composition and accompany it with per-family results. The
simulator queueing, grader dispatch, model configurations, and
reproducibility details are provided in the appendix.

\vspace{-4px}
\section{Measuring Intent Drift}
\vspace{-4px}
\label{sec:results_measurement}

We first test whether increasing the construction budget produces a
monotonic difficulty response. The calibration study uses
\texttt{gpt-4.1} for construction, \texttt{gemini-3.5-flash} as the
tested model, and \texttt{deepseek-v4-flash} as the user simulator and
rubric judge. The easy, medium, and hard strata use variant/decoy budgets
of 1, 3, and 5, respectively. Increasing this joint budget lowers both
mean score and \car{}
(Table~\ref{tab:appendix-difficulty-calibration}).
Because the additional edits also lengthen the dialogue, this calibration
exhibits a monotonic response to increasing joint stale-information load;
it does not estimate separate causal effects for variants and decoys.
\vspace{-4px}
\begin{table}[H]
\centering
\small
\caption{Main benchmark results on \gentest{} ($n=125$ per condition).
\idg{} is the clean--hard \car{} difference; brackets give paired 95\%
confidence intervals.}
\label{tab:main-benchmark}
\begingroup
\renewcommand{\arraystretch}{1.08}
\setlength{\tabcolsep}{6pt}
\begin{tabular}{@{}l
  S[table-format=1.3]
  S[table-format=1.3]
  S[table-format=1.3]
  S[table-format=1.3]
  l@{}}
\toprule
& \multicolumn{4}{c}{Full-credit \car{}} & \\
\cmidrule(lr){2-5}
Tested model & {Clean} & {Easy drift} & {Medium drift} & {Hard drift}
  & {Hard \idg{} [95\% CI]} \\
\midrule
Claude Opus 4.7  & 0.592 & 0.344 & 0.240 & 0.184 & $0.408\;[0.304,0.512]$ \\
DeepSeek V4 Pro  & 0.584 & 0.368 & 0.376 & 0.200 & $0.384\;[0.296,0.472]$ \\
Gemini 3.5 Flash & 0.632 & 0.384 & 0.320 & 0.264 & $0.368\;[0.280,0.456]$ \\
GLM 5.2          & 0.576 & 0.368 & 0.304 & 0.232 & $0.344\;[0.264,0.432]$ \\
GPT-5.4          & 0.608 & 0.376 & 0.264 & 0.224 & $0.384\;[0.304,0.472]$ \\
HY3              & 0.592 & 0.328 & 0.304 & 0.200 & $0.392\;[0.304,0.480]$ \\
Qwen3.6 Plus     & 0.640 & 0.448 & 0.312 & 0.192 & $0.448\;[0.360,0.536]$ \\
Qwen3.5-122B     & 0.592 & 0.288 & 0.240 & 0.232 & $0.360\;[0.280,0.448]$ \\
\bottomrule
\end{tabular}
\endgroup
\end{table}

The paired validation shows a positive \idg{} for every tested model
in the main benchmark (Table~\ref{tab:main-benchmark}) and at every
stratum. The loss generally grows with joint stale-information load;
DeepSeek V4 Pro is the only model whose medium point estimate is slightly
below its easy estimate. The raw score grid in
Table~\ref{tab:appendix-cross-model-validation} shows the same overall
degradation pattern without collapsing continuous task scores into
full-credit indicators. Separately, the turn-matched comparison in
Table~\ref{tab:appendix-len-matched-control} shows that adding turns
without changing the final intent has a much smaller effect than the
corresponding drift trajectories.
\vspace{-4px}
\subsection{Task-Family Responses and a Turn-Matched Control}

\label{ssec:length-control}

The aggregate trend is not driven by a single task family
(Table~\ref{tab:appendix-variant-decoy-family}). LiveCodeBench, GSM8K,
and HumanEval show the largest declines, while BFCL is an exception:
its score rises under the jointly constructed condition. SummHay remains
near a low-score floor. These heterogeneous responses motivate reporting
both aggregate and per-family results. Exact family scores appear in
Table~\ref{tab:appendix-variant-decoy-family}.

The case-micro aggregate gives larger families more weight. As a
composition-robust check, an equal-family macro average over the seven
task-family rows also decreases monotonically, from $.497$ (easy) to
$.433$ (medium) and $.385$ (hard). Thus, the headline trend does not
depend on weighting task families by their number of cases. Case-paired
\idg{} by task family is reported in
Table~\ref{tab:family-paired-idg}.

The calibration trajectories also lengthen the dialogue, from 13.2 turns
on average in easy to 37.1 in hard. We therefore construct a turn-matched
no-drift control on \gentest{}. It presents the final intent in the first
turn and uses neutral no-change confirmations thereafter.
The control and drift conditions have identical turn counts and comparable
user-message length. The control scores .734, close to the .778
single-turn condition, whereas the corresponding drift trajectories score
.469. The .265 gap shows that turn count alone does not explain the drift
loss.

This control is intentionally narrow: it removes revisions while matching
length, since matching competing values and their withdrawal would
reintroduce the stale-information mechanism being tested. It therefore
does not identify separate effects of wording, edit type, or variant
versus decoy; it only rules out generic long-dialogue degradation as the
explanation.

\vspace{-4px}
\subsection{Interactive Transfer}

The General track tests a static final answer after a simulated dialogue.
We therefore also evaluate \intertest{}, in which the agent acts in a
live tool-use environment. Qwen3.5-122B is the tested agent;
DeepSeek-V4-Flash is the user simulator and rubric judge. The main
condition does not restate the final intent during the closing turns;
we retain the original closing protocol, which does restate the final
intent, only as a diagnostic.

Table~\ref{tab:appendix-vitabench-external} shows that the no-restate
condition lowers rubric score by $0.103$ relative to clean, while
explicitly restating the final intent in the diagnostic condition masks
this gap. The binary \car{} remains unchanged across the non-oracle
conditions, so this evidence concerns partial satisfaction in the
interactive protocol.

\vspace{-4px}
\vspace{-4px}
\section{From History Compression to Explicit State Folding}

\label{sec:harness_failure_analysis}

We first ask whether existing history-management strategies are sufficient
to handle evolving intent. We compare two external harnesses on the same
frozen medium-difficulty \gentest{} trajectories, using Qwen3.5-122B as
the tested model and DeepSeek-V4-Flash as the user simulator and rubric
judge. Deep Agents applies rolling summarization, while OpenHarness uses
two-phase compaction. Under the same evaluator, temperature, and output
budget, Deep Agents scores 0.354 and OpenHarness scores 0.392, compared
with 0.367 for bare multi-turn execution
(Table~\ref{tab:stateforge-harness}, Panel B). Thus, the evaluated
history-management harnesses do not reliably recover the performance lost
under evolving intent. These results motivate a closer look at what history compression leaves
unresolved. A compressed history can still preserve both a superseded
value and its replacement, leaving the generator to determine which one
remains active while solving the downstream task.

Stale information can also persist indirectly. An obsolete intent item may
already have produced intermediate conclusions, plans, or choices. Even if
the original item is later removed, these dependent conclusions can remain
and continue to influence generation. A summary or compacted history may
preserve them because they appear locally consistent despite depending on
outdated intent. This motivates \emph{state folding}: resolving the edit history into an
explicit representation of the current intent before downstream task
solving. State folding removes not only superseded intent items but also
conclusions that depend on them, while preserving information that remains
valid after the edit. History management asks what information to retain;
state folding additionally asks what information is still valid.

% \stateforge{} instantiates this design by separating state update from
% downstream generation. After each user update, it maintains an explicit
% active state and supplies that state to the base agent before generation.
% Section~\ref{sec:stateforge_method} describes the harness and evaluates this
% decomposition against bare execution and the history-management baselines.
\vspace{-4px}
\section{StateForge: State Folding and Transfer}

\label{sec:stateforge_method}

\stateforge{} is a train-free harness that converts an evolving
conversation into an explicit active-state estimate. After each user
turn, a tracker folds observed intent edits into the current active
state. Deleting or replacing an intent item removes it from the state;
conclusions derived from that item are invalidated as well. The state
distinguishes directly stated intent items from derived conclusions so
that dependent conclusions can be removed when their supporting intent
changes.

\begin{figure*}[t]
\centering
\includegraphics[width=.92\textwidth]{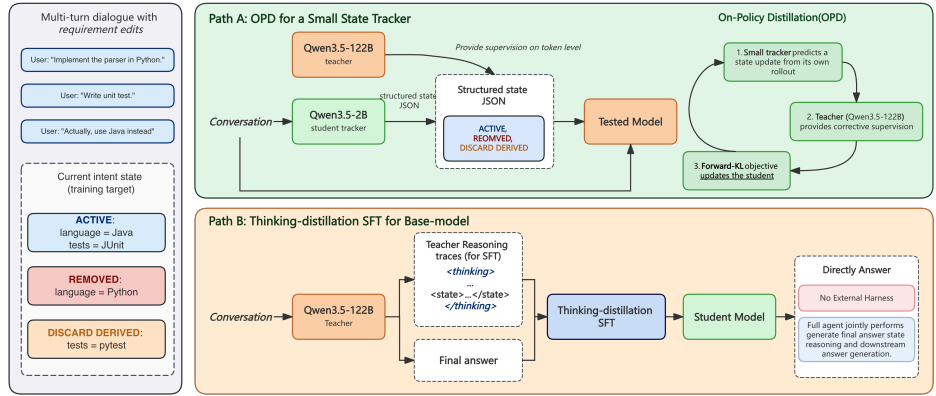}
\caption{\stateforge{} folds an evolving dialogue into an explicit active
state for external generation. The same state-update behavior supports
two transfer paths: on-policy distillation to a lightweight tracker and
thinking distillation to a standalone agent.}
\label{fig:stateforge-architecture}
\end{figure*}

For generation, the updated state is inserted between the cached dialogue
history and the current user input, keeping the prior conversation
cacheable while making active information recent at generation time. A
separate format tracker maintains output conventions, while a final gate
checks the generated answer against those conventions and can trigger
regeneration. The generator therefore receives an explicit active-state
estimate before solving the downstream task
(Figure~\ref{fig:stateforge-architecture}).

\begin{table}[t]
\vspace{-12pt}
\centering
\small
\caption{State folding on \gentest{}. StateForge improves over bare
multi-turn execution by $0.100$ mean score (95\% CI $[0.030,0.171]$).}
\label{tab:stateforge-harness}
\begingroup
\renewcommand{\arraystretch}{1.10}
\setlength{\tabcolsep}{9pt}
\begin{tabular}{lS[table-format=1.3]S[table-format=1.3]}
\toprule
Setting & {Mean score} & {\car{}} \\
\midrule
\multicolumn{3}{l}{\textit{Panel A: state-estimation analysis}} \\
Clean single-turn & 0.778 & {--} \\
Bare multi-turn & 0.367 & 0.264 \\
GT-oracle final state & 0.549 & {--} \\
\rowcolor{resultblue!8}\stateforge{} & {\bfseries 0.467} & {\bfseries 0.368} \\
\midrule
\multicolumn{3}{l}{\textit{Panel B: external-harness comparison}} \\
Deep Agents (rolling summary) & 0.354 & 0.256 \\
OpenHarness (compaction) & 0.392 & 0.296 \\
\rowcolor{resultblue!8}\stateforge{} & {\bfseries 0.467} & {\bfseries 0.368} \\
\bottomrule
\end{tabular}
\endgroup
\end{table}

\vspace{-4px}
\subsection{State Folding Repairs Intent Drift}

On \gentest{}, \stateforge{} raises mean score from $0.367$ for bare
multi-turn execution to $0.467$
(Table~\ref{tab:stateforge-harness}, Panel A). Supplying the
ground-truth final state raises the score further to $0.549$, showing
that state-estimation error remains material. This oracle supplies exact
final-state information at the final generation step while retaining the
full dialogue history; it is therefore not equivalent to the clean
single-turn condition. Its remaining $0.229$ gap to the clean score of
$0.778$ shows that errors remain even with exact state information under
the retained dialogue history. In the external-harness comparison, \stateforge{} also achieves higher
scores than the evaluated rolling-summary and compaction harnesses
(Panel B). All harness rows use Qwen3.5-122B as the tested model and
DeepSeek-V4-Flash as the simulator and rubric judge. Mean score includes
the continuous data-to-text and summary metrics; \car{} is the full-credit
rate over the complete test set.

\vspace{-0px}
\subsection{Tracker Capacity and Transfer}

The ground-truth-state result shows remaining headroom in final-state
estimation. We next ask whether this tracking role requires a model as
large as the base agent. In the tracker-scale comparison, 2B and 4B
trackers are significantly below the 122B reference, whereas the 9B
tracker is statistically indistinguishable from it. We therefore use the
9B result as evidence that the modular state-estimation role does not
require a tracker at base-agent scale.

We next train the tracker sub-role while keeping the 122B base agent
fixed. One \opd{} iteration raises a 2B tracker from $0.266$ to $0.445$
on \gentest{}, while 122B tracker scores $0.428$ under the same
evaluation protocol. Training pool is disjoint from \gentest{}.
Because the generator remains frozen, this improvement isolates adaptation
of the tracker within the modular stack.

Finally, in a separate training setting, we explore whether state-folding
behavior can be internalized into the base agent. Thinking distillation
raises a standalone 35B agent from $0.296$ to $0.461$ on the \gentest{}
drift condition without an external harness. This is not a direct system
comparison with the OPD tracker stack; rather, it tests a different
deployment in which state maintenance is absorbed into the agent itself.

\begin{figure*}[t]
\centering
\begin{minipage}[t]{0.58\textwidth}
\vspace{0pt}
\centering
\begin{tikzpicture}
\begin{axis}[
  width=0.96\linewidth,
  height=5.25cm,
  title={(a) Tracker capacity},
  title style={font=\small\bfseries},
  xmode=log,
  log basis x=10,
  xmin=1.6,
  xmax=150,
  xtick={2,4,9,27,35,122},
  xticklabels={2B,4B,9B,27B,35B,122B},
  xlabel={Tracker parameters (log scale)},
  ymin=0.08,
  ymax=0.48,
  ytick={0.1,0.2,0.3,0.4},
  ylabel={Score},
  tick label style={font=\scriptsize},
  x tick label style={font=\tiny,rotate=55,anchor=east},
  label style={font=\scriptsize},
  grid=major,
  grid style={black!10},
  axis line style={black!55},
  legend style={font=\scriptsize,draw=none,fill=none,
    at={(0.5,0.02)},anchor=south,legend columns=2},
]
\addplot+[resultblue,very thick,mark=*,mark options={solid}] coordinates {
  (2,0.156) (4,0.312) (9,0.399) (27,0.366) (35,0.395) (122,0.428)
};
\addlegendentry{Mean score}
\addplot+[resultorange,very thick,dashed,mark=square*,mark options={solid}] coordinates {
  (2,0.112) (4,0.240) (9,0.312) (27,0.272) (35,0.304) (122,0.344)
};
\addlegendentry{Full-credit CAR}
\end{axis}
\end{tikzpicture}
\end{minipage}\hfill
\begin{minipage}[t]{0.38\textwidth}
\vspace{0pt}
\centering
\begin{tikzpicture}
\begin{axis}[
  width=0.96\linewidth,
  height=2.55cm,
  title={\shortstack{(b) OPD tracker adaptation\\[-1pt]
    \tiny Frozen 122B agent + 2B tracker}},
  title style={font=\scriptsize\bfseries},
  xmin=0.20,
  xmax=0.50,
  xtick={0.2,0.3,0.4,0.5},
  ymin=-0.20,
  ymax=0.20,
  ytick=\empty,
  xlabel={Held-out full-episode score},
  tick label style={font=\tiny},
  label style={font=\tiny},
  axis y line=none,
  axis x line*=bottom,
  xmajorgrids,
  grid style={black!10},
  clip=false,
]
\addplot+[black!55,very thick,->] coordinates {(0.266,0) (0.445,0)};
\addplot+[only marks,color=black!55,mark=*,mark size=2.7pt,
  mark options={fill=black!28}]
  coordinates {(0.266,0)};
\addplot+[only marks,color=resultteal,mark=*,mark size=2.7pt,
  mark options={fill=resultteal}]
  coordinates {(0.445,0)};
\node[font=\tiny,anchor=south] at (axis cs:0.266,0.025) {Cold-start};
\node[font=\tiny,anchor=north] at (axis cs:0.266,-0.025) {0.266};
\node[font=\tiny,anchor=south] at (axis cs:0.445,0.025) {After OPD};
\node[font=\tiny,anchor=north] at (axis cs:0.445,-0.025) {0.445};
\end{axis}
\end{tikzpicture}

\vspace{2pt}

\begin{tikzpicture}
\begin{axis}[
  width=0.96\linewidth,
  height=2.55cm,
  title={\shortstack{(c) Full-agent adaptation\\[-1pt]
    \tiny Trained 35B agent; no external tracker}},
  title style={font=\scriptsize\bfseries},
  xmin=0.20,
  xmax=0.50,
  xtick={0.2,0.3,0.4,0.5},
  ymin=-0.20,
  ymax=0.20,
  ytick=\empty,
  xlabel={\gentest{} mean score},
  tick label style={font=\tiny},
  label style={font=\tiny},
  axis y line=none,
  axis x line*=bottom,
  xmajorgrids,
  grid style={black!10},
  clip=false,
]
\addplot+[black!55,very thick,->] coordinates {(0.296,0) (0.461,0)};
\addplot+[only marks,color=black!55,mark=*,mark size=2.7pt,
  mark options={fill=black!28}]
  coordinates {(0.296,0)};
\addplot+[only marks,color=resultteal,mark=*,mark size=2.7pt,
  mark options={fill=resultteal}]
  coordinates {(0.461,0)};
\node[font=\tiny,anchor=south] at (axis cs:0.296,0.025) {Base};
\node[font=\tiny,anchor=north] at (axis cs:0.296,-0.025) {0.296};
\node[font=\tiny,anchor=south] at (axis cs:0.461,0.025) {Distilled};
\node[font=\tiny,anchor=north] at (axis cs:0.461,-0.025) {0.461};
\end{axis}
\end{tikzpicture}
\end{minipage}
\caption{Tracker capacity and two distinct adaptation settings. (a) Tracker
results are positioned by parameter count on a logarithmic axis. (b) OPD
updates a 2B tracker paired with a frozen 122B base agent; values are
held-out full-episode scores. (c) Thinking distillation updates a standalone
35B agent with no external tracker; values are \gentest{} mean scores.
Panels (b) and (c) use different system configurations and evaluation
protocols, so only the before--after change within each panel should be
compared. Tracker-scale confidence intervals appear in
Appendix~\ref{tab:tracker-scale}.}
\label{fig:stateforge-transfer}
\vspace{-12px}
\end{figure*}
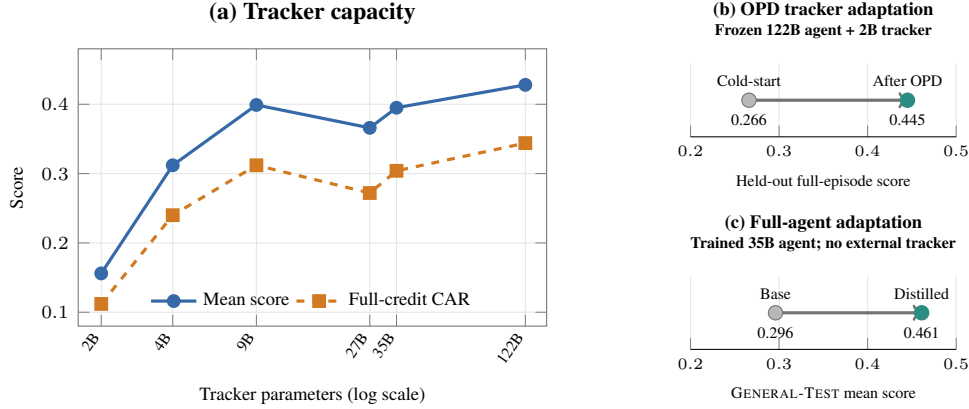

\vspace{-4px}
\subsection{From State Tracking to Agent Behavior}
\vspace{-4px}
The transfer experiments target two distinct deployment choices. In the
first, the generator remains unchanged and a smaller tracker supplies the
active state at inference time. This separates state maintenance from
downstream task solving: the tracker determines which intent items remain
active after an edit, while the base agent continues to solve the source
task. The scale comparison and OPD result show that this role can be
performed by a substantially smaller model and improved through on-policy
correction.

In the second deployment, the external tracker is removed. Thinking
distillation exposes a 35B agent to teacher rollouts that make state
updates explicit in its reasoning, then evaluates the trained agent on
the \gentest{} drift condition without a harness. The result does not
establish that all state-folding operations have been learned or that the
learned behavior generalizes to every update regime. The improvement is
also accompanied by a modest decrease on the clean condition: the
distilled model scores $.718$, compared with $.745$ for the base model.

These two paths answer complementary questions. The tracker path asks
whether current intent can be maintained efficiently as a modular
operation; the full-agent path asks whether the same behavior can be
absorbed into the model itself. Keeping the two settings separate makes
their training and deployment assumptions explicit and enables future
work to compare modular and learned state maintenance under a shared
evaluation protocol.

\vspace{-4px}
\subsection{What the Ablation Establishes}

The system-level comparison above does not determine whether the gain
comes specifically from the explicit state representation or from other
components of the harness. We therefore compare the complete harness with
gate-free state injection and two recap baselines matched to the full
harness in model calls and token budget under a separate
component-matched re-run. 
At $n=125$, the paired differences between each recap baseline and the
full harness have 95\% confidence intervals that include zero, so the
current experiment does not establish representation superiority over a
matched recap (Table~\ref{tab:appendix-component-ablation}). The oracle
condition, by contrast, shows a positive paired difference from the full
harness, demonstrating material headroom from exact state estimation
within the same history-preserving interface. This result does not
identify the tracker as the dominant source of the clean--drift gap.
Detailed tracker-scale results and training progression are reported in
the appendix.
\vspace{-8px}
\section{Analysis, Limitations, and Broader Impact}

\label{sec:analysis}
\vspace{-4px}
\subsection{What the Benchmark Measures}

The variant/decoy calibration provides a controllable stress condition.
Its $b\in\{1,3,5\}$ schedule jointly introduces more superseded
alternatives and more withdrawn constraints, while also lengthening the
edit sequence. The simulator does not close by restating the final intent.
The resulting score decline therefore characterizes a joint
stale-information stress condition rather than the separate causal
effects of variants, decoys, or length. The intent-drift claim does not
rest on this slope alone: case-paired clean--drift gaps remain positive
across models and strata, and the turn-matched control shows substantially
less degradation than the drift condition. Per-family results further
show heterogeneous sensitivity to the construction: code and math decline
sharply, whereas BFCL improves under the same setting.

Trace inspection suggests that intent drift is not limited to explicit
retention of deleted content. In drift cases containing a
\textsc{Delete} edit, agents often omit the deleted goal from the final
answer. In BFCL, a common \textsc{Replace} failure instead binds the new
value to the wrong function-call slot. These observations motivate state
tracking that represents the identity and dependencies of intent items,
rather than relying only on retrieval over prior utterances.
\vspace{-4px}
\subsection{Simulator and Evaluator Validity}

The user simulator is an LLM, so its wording is part of the evaluation
protocol. Diversified utterance prompts, explicit final-answer triggers,
and per-goal edit queues constrain how scheduled edits are realized. In
a sampled audit of 100 edit-phase utterances from the DeepSeek-V4-Flash
simulator used throughout our experiments, fidelity reaches 96\% and
coherence reaches 97\%. The audit indicates that invented and
contradictory intent changes are uncommon in the sampled dialogues.

The interactive VitaBench study exposes a related protocol issue. The
main interactive condition does not restate the final intent. In a
separate diagnostic condition, supplying the complete final intent during
the closing turns removes the measured partial-credit drift gap. A
benchmark can therefore inadvertently supply the state that it intends
to test; the restating condition is not used as a difficulty factor.
\vspace{-4px}
\subsection{Limitations}

\bench{} represents current intent as a finite set of atomic goals and
therefore targets tasks with executable or decomposable intent structure.
The calibration couples variant and decoy budgets, supporting a joint
stale-load effect rather than separate marginal effects; the turn-matched
control rules out turn count alone, but message ordering and edit wording
remain properties of the drift construction. The external-harness
comparison and \intertest{} each use a single tested model (Qwen3.5-122B),
the former on replayed trajectories. The component-matched ablation does
not isolate the explicit state representation as the sole source of the
system-level gain, and the scaling behavior of both transfer paths remains
open.

% \subsection{Broader Impact}

% Following withdrawn intent could pose deployment risks in domains such as
% medical, legal, and financial assistance. The benchmark uses public
% research datasets, collects no human-subject data, and releases only
% synthetic, task-focused interaction content.
\vspace{-4px}
\section{Conclusion}
\label{sec:conclusions}

\bench{} measures whether agents follow current user intent after earlier
information is superseded or withdrawn; its construction makes obsolete
intent grader-consequential and yields a controlled difficulty response.
The evaluated history-management harnesses leave substantial errors, while
\stateforge{}, which maintains an explicit active state, improves
performance. This tracking role does not require a base-agent-scale model:
\opd{} further improves a small tracker, and thinking distillation begins
to internalize the behavior into the agent. Together, these results frame
intent drift as a measurable state-maintenance problem that modular and
learned approaches can partially mitigate.

% \subsubsection*{AI use statement}

% Generative AI tools assisted code scaffolding for evaluation and plotting
% and language editing of portions of the manuscript. The authors reviewed,
% executed, and verified AI-assisted code and reviewed every edited passage.
% The LLM-based construction, user simulation, and judging components are
% part of the evaluated protocol, not assistance used to prepare the paper.
% The authors take responsibility for the final text, artifacts, and claims.

% \subsubsection*{Ethics statement}

% All tracks are constructed from public research benchmarks under their
% original terms; no human-subject data or personal information is collected.
% Acting on withdrawn requirements can cause harm in high-stakes assistant
% settings. The released benchmark content is synthetic and task-focused.

% \subsubsection*{Reproducibility statement}

% The appendix specifies benchmark construction, evaluation, graders, and
% model configurations. The anonymized supplement contains the harness,
% construction and evaluation pipelines, frozen \gentest{} data, and archived
% per-case traces for the reported frozen-protocol runs, with aggregation
% scripts for reported aggregates. Commercial-API traces are not
% redistributable; the corresponding evaluation pipeline is included.

\bibliographystyle{iclr2027_conference}
\bibliography{iclr2027_conference}

@inproceedings{laban2025lost,
  author = {Philippe Laban and Hiroaki Hayashi and Yingbo Zhou and Jennifer Neville},
  title = {LLMs Get Lost In Multi-Turn Conversation},
  booktitle = {Proceedings of the International Conference on Learning Representations (ICLR)},
  year = {2025},
}

@inproceedings{sirdeshmukh2025multichallenge,
  author = {Ved Sirdeshmukh and Kaustubh Deshpande and Johannes Mols and Lifeng Jin and Ed-Yeremai Cardona and Dean Lee and Jeremy Kritz and Willow Primack and Summer Yue and Chen Xing},
  title = {MultiChallenge: A Realistic Multi-Turn Conversation Evaluation Benchmark Challenging to Frontier LLMs},
  booktitle = {Proceedings of the Annual Meeting of the Association for Computational Linguistics (ACL)},
  year = {2025},
}

@article{liu2023lost,
  author = {Nelson F. Liu and Kevin Lin and John Hewitt and Ashwin Paranjape and Michele Bevilacqua and Fabio Petroni and Percy Liang},
  title = {Lost in the Middle: How Language Models Use Long Contexts},
  journal = {Transactions of the Association for Computational Linguistics (TACL)},
  year = {2023},
}

@inproceedings{kwan2024mteval,
  author = {Wai-Chung Kwan and Xingshan Zeng and Yuxin Jiang and Yufei Wang and Liangyou Li and Lifeng Shang and Xin Jiang and Qun Liu and Kam-Fai Wong},
  title = {MT-Eval: A Multi-Turn Capabilities Evaluation Benchmark for Large Language Models},
  booktitle = {Proceedings of the Conference on Empirical Methods in Natural Language Processing (EMNLP)},
  year = {2024},
}

@article{katsis2025mtrag,
  author = {Yannis Katsis and Sara Rosenthal and Kshitij Fadnis and Chulaka Gunasekara and Young-Suk Lee and Lucian Popa and Vraj Shah and Huaiyu Zhu and Danish Contractor and Marina Danilevsky},
  title = {MTRAG: A Multi-Turn Conversational Benchmark for Evaluating Retrieval-Augmented Generation Systems},
  journal = {Transactions of the Association for Computational Linguistics (TACL)},
  year = {2025},
}

@inproceedings{zhang2025ace,
  author = {Qizheng Zhang and Changran Hu and Shubhangi Upasani and Boyuan Ma and Fenglu Hong and Vamsidhar Kamanuru and Jay Rainton and Chen Wu and Mengmeng Ji and Hanchen Li and Urmish Thakker and James Zou and Kunle Olukotun},
  title = {Agentic Context Engineering: Evolving Contexts for Self-Improving Language Models},
  booktitle = {Proceedings of the International Conference on Learning Representations (ICLR)},
  year = {2025},
}

@inproceedings{lee2024readagent,
  author = {Kuang-Huei Lee and Xinyun Chen and Hiroki Furuta and John Canny and Ian Fischer},
  title = {A Human-Inspired Reading Agent with Gist Memory of Very Long Contexts},
  booktitle = {Proceedings of the International Conference on Machine Learning (ICML)},
  year = {2024},
}

@inproceedings{jiang2023longllmlingua,
  author = {Huiqiang Jiang and Qianhui Wu and Xufang Luo and Dongsheng Li and Chin-Yew Lin and Yuqing Yang and Lili Qiu},
  title = {LongLLMLingua: Accelerating and Enhancing LLMs in Long Context Scenarios via Prompt Compression},
  booktitle = {Proceedings of the Annual Meeting of the Association for Computational Linguistics (ACL)},
  year = {2023},
}

@inproceedings{agarwal2023gkd,
  author = {Rishabh Agarwal and Nino Vieillard and Yongchao Zhou and Piotr Stanczyk and Sabela Ramos and Matthieu Geist and Olivier Bachem},
  title = {On-Policy Distillation of Language Models: Learning from Self-Generated Mistakes},
  booktitle = {Proceedings of the International Conference on Learning Representations (ICLR)},
  year = {2023},
}

@inproceedings{hsieh2023distilling,
  author = {Cheng-Yu Hsieh and Chun-Liang Li and Chih-Kuan Yeh and Hootan Nakhost and Yasuhisa Fujii and Alexander Ratner and Ranjay Krishna and Chen-Yu Lee and Tomas Pfister},
  title = {Distilling Step-by-Step! Outperforming Larger Language Models with Less Training Data and Smaller Model Sizes},
  booktitle = {Findings of the Association for Computational Linguistics (ACL)},
  year = {2023},
}

@misc{mukherjee2023orca,
  author = {Subhabrata Mukherjee and Arindam Mitra and Ganesh Jawahar and Sahaj Agarwal and Hamid Palangi and Ahmed Awadallah},
  title = {Orca: Progressive Learning from Complex Explanation Traces of GPT-4},
  year = {2023},
  eprint = {2306.02707},
  archivePrefix = {arXiv},
  primaryClass = {cs.CL},
  url = {https://arxiv.org/abs/2306.02707},
}

@inproceedings{zhu2024beliefs,
  author = {Wentao Zhu and Zhining Zhang and Yizhou Wang},
  title = {Language Models Represent Beliefs of Self and Others},
  booktitle = {Proceedings of the International Conference on Machine Learning (ICML)},
  year = {2024},
}

@misc{guan2024deliberative,
  author = {Melody Y. Guan and Manas Joglekar and Eric Wallace and Saachi Jain and Boaz Barak and Alec Helyar and Rachel Dias and Andrea Vallone and Hongyu Ren and Jason Wei and Hyung Won Chung and Sam Toyer and Johannes Heidecke and Alex Beutel and Amelia Glaese},
  title = {Deliberative Alignment: Reasoning Enables Safer Language Models},
  year = {2024},
  eprint = {2412.16339},
  archivePrefix = {arXiv},
  primaryClass = {cs.CL},
  url = {https://arxiv.org/abs/2412.16339},
}

@misc{chen2021humaneval,
  author = {Mark Chen and Jerry Tworek and Heewoo Jun and Qiming Yuan and Henrique Ponde de Oliveira Pinto and Jared Kaplan and Harri Edwards and Yuri Burda and Nicholas Joseph and Greg Brockman and Alex Ray and Raul Puri and Gretchen Krueger and Michael Petrov and Heidy Khlaaf and Girish Sastry and Pamela Mishkin and Brooke Chan and Scott Gray and Nick Ryder and Mikhail Pavlov and Alethea Power and Lukasz Kaiser and Mohammad Bavarian and Clemens Winter and Philippe Tillet and Felipe Petroski Such and Dave Cummings and Matthias Plappert and Fotios Chantzis and Elizabeth Barnes and Ariel Herbert-Voss and William Hebgen Guss and Alex Nichol and Alex Paino and Nikolas Tezak and Jie Tang and Igor Babuschkin and Suchir Balaji and Shantanu Jain and William Saunders and Christopher Hesse and Andrew N. Carr and Jan Leike and Josh Achiam and Vedant Misra and Evan Morikawa and Alec Radford and Matthew Knight and Miles Brundage and Mira Murati and Katie Mayer and Peter Welinder and Bob McGrew and Dario Amodei and Sam McCandlish and Ilya Sutskever and Wojciech Zaremba},
  title = {Evaluating Large Language Models Trained on Code},
  year = {2021},
  eprint = {2107.03374},
  archivePrefix = {arXiv},
  primaryClass = {cs.CL},
  url = {https://arxiv.org/abs/2107.03374},
}

@misc{cobbe2021gsm8k,
  author = {Karl Cobbe and Vineet Kosaraju and Mohammad Bavarian and Mark Chen and Heewoo Jun and Lukasz Kaiser and Matthias Plappert and Jerry Tworek and Jacob Hilton and Reiichiro Nakano and Christopher Hesse and John Schulman},
  title = {Training Verifiers to Solve Math Word Problems},
  year = {2021},
  eprint = {2110.14168},
  archivePrefix = {arXiv},
  primaryClass = {cs.CL},
  url = {https://arxiv.org/abs/2110.14168},
}

@inproceedings{yu2018spider,
  author = {Tao Yu and Rui Zhang and Kai Yang and Michihiro Yasunaga and Dongxu Wang and Zifan Li and James Ma and Irene Li and Qingning Yao and Shanelle Roman and Zilin Zhang and Dragomir Radev},
  title = {Spider: A Large-Scale Human-Labeled Dataset for Complex and Cross-Domain Semantic Parsing and Text-to-SQL Task},
  booktitle = {Proceedings of the Conference on Empirical Methods in Natural Language Processing (EMNLP)},
  year = {2018},
}

@inproceedings{parikh2020totto,
  author = {Ankur P. Parikh and Xuezhi Wang and Sebastian Gehrmann and Manaal Faruqui and Bhuwan Dhingra and Diyi Yang and Dipanjan Das},
  title = {ToTTo: A Controlled Table-To-Text Generation Dataset},
  booktitle = {Proceedings of the Conference on Empirical Methods in Natural Language Processing (EMNLP)},
  year = {2020},
}

@misc{laban2024summhay,
  author = {Philippe Laban and Alexander R. Fabbri and Caiming Xiong and Chien-Sheng Wu},
  title = {Summary of a Haystack: A Challenge to Long-Context LLMs and RAG Systems},
  year = {2024},
  eprint = {2407.01370},
  archivePrefix = {arXiv},
  primaryClass = {cs.CL},
  url = {https://arxiv.org/abs/2407.01370},
}

@inproceedings{jain2024livecodebench,
  author = {Naman Jain and King Han and Alex Gu and Wen-Ding Li and Fanjia Yan and Tianjun Zhang and Sida Wang and Armando Solar-Lezama and Koushik Sen and Ion Stoica},
  title = {LiveCodeBench: Holistic and Contamination Free Evaluation of Large Language Models for Code},
  booktitle = {Proceedings of the International Conference on Learning Representations (ICLR)},
  year = {2024},
}

@inproceedings{he2025vitabench,
  author = {Wei He and Yueqing Sun and Hongyan Hao and Xueyuan Hao and Zhikang Xia and Qi Gu and Chengcheng Han and Dengchang Zhao and Hui Su and Kefeng Zhang and Man Gao and Xi Su and Xiaodong Cai and Xunliang Cai and Yu Yang and Yunke Zhao},
  title = {VitaBench: Benchmarking LLM Agents with Versatile Interactive Tasks in Real-world Applications},
  booktitle = {Proceedings of the International Conference on Learning Representations (ICLR)},
  year = {2025},
}

@inproceedings{kim2023entity,
  author = {Najoung Kim and Sebastian Schuster},
  title = {Entity Tracking in Language Models},
  booktitle = {Proceedings of the Annual Meeting of the Association for Computational Linguistics (ACL)},
  year = {2023},
  url = {https://aclanthology.org/2023.acl-long.558/},
}

@inproceedings{patil2025bfcl,
  author = {Shishir G. Patil and Huanzhi Mao and Charlie Cheng-Jie Ji and Fanjia Yan and Vishnu Suresh and Ion Stoica and Joseph E. Gonzalez},
  title = {The Berkeley Function Calling Leaderboard (BFCL): From Tool Use to Agentic Evaluation of Large Language Models},
  booktitle = {Proceedings of the International Conference on Machine Learning (ICML)},
  year = {2025},
  url = {https://proceedings.mlr.press/v267/patil25a.html},
}

@misc{tack2026evolvingintent,
  author = {Jihoon Tack and Philippe Laban and Jennifer Neville},
  title = {LLMs Get Lost in Evolving User Intent},
  year = {2026},
  eprint = {2607.20734},
  archivePrefix = {arXiv},
  primaryClass = {cs.LG},
  url = {https://arxiv.org/abs/2607.20734},
}

@misc{zou2026interruptbench,
  author = {Henry Peng Zou and Chunyu Miao and Wei-Chieh Huang and Yankai Chen and Yue Zhou and Hanrong Zhang and Yaozu Wu and Liancheng Fang and Zhengyao Gu and Zhen Zhang and Kening Zheng and Fangxin Wang and Yi Nian and Shanghao Li and Wenzhe Fan and Langzhou He and Weizhi Zhang and Xue Liu and Philip S. Yu},
  title = {When Users Change Their Mind: Evaluating Interruptible Agents in Long-Horizon Web Navigation},
  year = {2026},
  eprint = {2604.00892},
  archivePrefix = {arXiv},
  primaryClass = {cs.CL},
  url = {https://arxiv.org/abs/2604.00892},
}

@inproceedings{jia2026evolif,
  author = {Qi Jia and Ye Shen and Xiujie Song and Kaiwei Zhang and Shibo Wang and Dun Pei and Xiangyang Zhu and Guangtao Zhai},
  title = {One Battle After Another: Probing {LLMs'} Limits on Multi-Turn Instruction Following with a Benchmark Evolving Framework},
  booktitle = {Proceedings of the Annual Meeting of the Association for Computational Linguistics (ACL)},
  year = {2026},
  url = {https://aclanthology.org/2026.acl-long.433/},
}

@inproceedings{uddin2026memora,
  author = {Md Nayem Uddin and Kumar Shubham and Eduardo Blanco and Chitta Baral and Gengyu Wang},
  title = {From Recall to Forgetting: Benchmarking Long-Term Memory for Personalized Agents},
  booktitle = {Findings of the Association for Computational Linguistics (ACL)},
  year = {2026},
  url = {https://aclanthology.org/2026.findings-acl.1337/},
}

@misc{patel2026supersede,
  author = {Vedant Patel},
  title = {Supersede: Diagnosing and Training the Memory-Update Gap in {LLM} Agents},
  year = {2026},
  eprint = {2606.27472},
  archivePrefix = {arXiv},
  primaryClass = {cs.CL},
  url = {https://arxiv.org/abs/2606.27472},
}

% !TEX root = ../iclr2027_conference.tex
\appendix

\Needspace{8\baselineskip}
\section{Benchmark Construction and Evaluation Details}

\begin{table}[H]
\centering
\small
\caption{\bench{} source sets and paper strata.}
\label{tab:source-sets}
\begin{tabular}{lrl}
\toprule
Stratum & Cases & Role \\
\midrule
General source tasks & --- & code, math, SQL, data-to-text, summary, tool use \\
General calibration pool & 627 & variant/decoy stress condition \\
\gentest{} & 125 & cross-model, control, and harness evaluation \\
\intertest{} & 100 & interactive transfer \\
Training pool & 502 & OPD and thinking-distillation source pool \\
\bottomrule
\end{tabular}
\end{table}

Each source problem is decomposed into ordered atomic shards. A shard is
atomic when changing it changes the final action. We construct
behavior-changing variants and plausible decoys, reject pure rewordings,
and enforce dependency constraints when sampling edit plans. Human
annotators inspect every variant, decoy, and semantic-dependency triple.
A downstream audit of \gentest{} recovers the annotated
$\gstar$ for 95\% of cases; further quality-control details appear in
Section~\ref{app:qc-rubrics}.

\begin{table}[H]
\centering
\small
\caption{Task-native graders by task family.}
\label{tab:graders}
\begin{tabular}{ll}
\toprule
Task family & Grader \\
\midrule
code & final function against public and hidden tests \\
math & extracted numeric answer \\
database & SQL execution against Spider databases \\
data-to-text & sacreBLEU \\
summary & SummHay LLM-based grader \\
actions & BFCL AST check \\
VitaBench & domain-specific evaluator \\
\bottomrule
\end{tabular}
\end{table}

The simulator realizes per-goal edit queues and a decoy queue, preserving
within-goal order while interleaving feasible edits. Every expected
\gentest{} case enters reported aggregates; absent scorable outputs count
as zero. The primary tested backbone for the single-model analyses is
Qwen3.5-122B-A10B-FP8 served locally with vLLM. DeepSeek-V4-Flash serves
as the user simulator and rubric judge throughout; complete model,
serving, seed, and bootstrap details are supplied in the anonymized
artifact.

\Needspace{8\baselineskip}
\section{Detailed Benchmark Results}

\begin{table}[H]
\centering
\small
\caption{Aggregate response in the joint variant/decoy difficulty calibration.}
\label{tab:appendix-difficulty-calibration}
\begin{tabular}{lcccc}
\toprule
Stratum & Realized variants & Realized decoys & Mean score & \car{} \\
\midrule
Easy & 0.95 & 0.61 & 0.476 & 0.381 \\
Medium & 1.71 & 1.67 & 0.419 & 0.327 \\
Hard & 2.24 & 2.50 & 0.384 & 0.298 \\
\midrule
$\Delta$(easy$\to$hard) & $+1.29$ & $+1.89$ & $-0.092$ & $-0.083$ \\
\bottomrule
\end{tabular}
\end{table}

\begin{table}[H]
\centering
\small
\caption{Case-paired clean--drift \idg{} by tested model and difficulty.
Brackets give paired 95\% confidence intervals.}
\label{tab:appendix-model-idg}
\begin{tabular}{lcccc}
\toprule
Tested model & Clean \car{} & Easy & Medium & Hard \\
\midrule
Claude Opus 4.7 & .592 & .248 [.160,.336] & .352 [.256,.448] & .408 [.304,.512] \\
DeepSeek V4 Pro & .584 & .216 [.128,.304] & .208 [.128,.288] & .384 [.296,.472] \\
Gemini 3.5 Flash & .632 & .248 [.168,.336] & .312 [.232,.400] & .368 [.280,.456] \\
GLM 5.2 & .576 & .208 [.128,.288] & .272 [.192,.352] & .344 [.264,.432] \\
GPT-5.4 & .608 & .232 [.160,.312] & .344 [.256,.432] & .384 [.304,.472] \\
HY3 & .592 & .264 [.176,.352] & .288 [.200,.384] & .392 [.304,.480] \\
Qwen3.6 Plus & .640 & .192 [.120,.264] & .328 [.248,.416] & .448 [.360,.536] \\
Qwen3.5-122B & .592 & .304 [.224,.384] & .352 [.264,.440] & .360 [.280,.448] \\
\bottomrule
\end{tabular}
\end{table}

\begin{table}[H]
\centering
\small
\caption{Per-family response to increasing joint variant/decoy load.}
\label{tab:appendix-variant-decoy-family}
\begin{tabular}{lrcccc}
\toprule
Task family & $n$ & easy & medium & hard & $\Delta$(easy$\to$hard) \\
\midrule
code (LiveCodeBench) & 55 & .582 & .418 & .273 & $-.309$ \\
math (GSM8K) & 103 & .592 & .476 & .408 & $-.184$ \\
code (HumanEval) & 45 & .689 & .622 & .511 & $-.178$ \\
database (Spider) & 107 & .430 & .327 & .308 & $-.121$ \\
data-to-text (ToTTo) & 120 & .367 & .352 & .325 & $-.042$ \\
summary (SummHay) & 92 & .165 & .169 & .163 & $-.003$ \\
actions (BFCL) & 105 & .657 & .667 & .705 & $+.048$ \\
\midrule
\textbf{ALL} & \textbf{627} & \textbf{.476} & \textbf{.419} & \textbf{.384} & \textbf{$-.092$} \\
\bottomrule
\end{tabular}
\end{table}

\begin{table}[H]
\centering
\small
\caption{Raw cross-model validation on the \gentest{} stress grid. Each
cell is case-micro mean task-native score / full-credit \car{}.}
\label{tab:appendix-cross-model-validation}
\begin{tabular}{lrrrr}
\toprule
Tested model & Easy & Medium & Hard & Avg. \\
\midrule
Claude Opus 4.7 & .444/.344 & .326/.240 & .268/.184 & .346/.256 \\
DeepSeek V4 Pro & .462/.368 & .468/.376 & .277/.200 & .402/.315 \\
Gemini 3.5 Flash & .482/.384 & .416/.320 & .360/.264 & .420/.323 \\
GLM 5.2 & .458/.368 & .404/.304 & .300/.232 & .387/.301 \\
GPT-5.4 & .479/.376 & .338/.264 & .302/.224 & .373/.288 \\
HY3 & .400/.328 & .368/.304 & .257/.200 & .342/.277 \\
Qwen3.6 Plus & .569/.448 & .408/.312 & .274/.192 & .417/.317 \\
Qwen3.5-122B & .288/.288 & .240/.240 & .232/.232 & .253/.253 \\
\bottomrule
\end{tabular}
\end{table}

\begin{table}[H]
\centering
\small
\caption{Case-paired full-credit \idg{} by task family on \gentest{}
(Gemini 3.5 Flash). ToTTo and SummHay have zero \idg{} because neither
condition attains full credit; their partial-credit changes are represented
in mean score rather than \car{}.}
\label{tab:family-paired-idg}
\begin{tabular}{lrccc}
\toprule
Task family & $n$ & Easy & Medium & Hard \\
\midrule
code (LiveCodeBench) & 10 & .700 & .400 & .900 \\
code (HumanEval) & 10 & .400 & .500 & .500 \\
math (GSM8K) & 21 & .286 & .381 & .429 \\
database (Spider) & 21 & .381 & .619 & .762 \\
data-to-text (ToTTo) & 24 & .000 & .000 & .000 \\
summary (SummHay) & 18 & .000 & .000 & .000 \\
actions (BFCL) & 21 & .286 & .429 & .333 \\
\bottomrule
\end{tabular}
\end{table}

\Needspace{8\baselineskip}
\section{Controls and Interactive Transfer}
\label{app:anchor-check}

The difficulty calibration uses variant/decoy budgets of 1, 3, and 5 for the easy, medium,
and hard strata. These budgets jointly increase the available same-slot
alternatives and eligible withdrawn decoys. They also increase mean
dialogue length from 13.2 to 37.1 turns. This coupling is why the main text
interprets the calibration as a joint stale-information stress condition and
uses a separate control for length. The final-answer trigger does not
restate $\gstar$ in any of the three strata.

We additionally construct a turn-matched no-drift control on \gentest{}.
It states the final requirements in the first turn
and uses no-change confirmations thereafter. The control and drift
conditions have identical turn counts and comparable user-message length
(13.2k versus 14.6k mean characters). The control is much closer to the
single-turn condition than the drift trajectories on most families,
whereas GSM8K exhibits a separate premature-commitment failure. This
control targets the length hypothesis: it is not designed to reproduce
the obsolete competing values, because doing so would reintroduce the
drift treatment. It therefore does not separately identify effects of
message order, edit wording, variants, and decoys.

\begin{table}[H]
\centering
\small
\caption{Turn-matched no-drift control on \gentest{}
(Qwen3.5-122B, mean score).}
\label{tab:appendix-len-matched-control}
\begin{tabular}{lrcccc}
\toprule
Task family & $n$ & single-turn & matched control & drift & control$-$drift \\
\midrule
code (LiveCodeBench) & 10 & 1.000 & 1.000 & .600 & +.400 \\
math (GSM8K) & 21 & .952 & .667 & .667 & +.000 \\
code (HumanEval) & 10 & 1.000 & 1.000 & .400 & +.600 \\
database (Spider) & 21 & .952 & .952 & .571 & +.381 \\
data-to-text (ToTTo) & 24 & .499 & .497 & .219 & +.278 \\
summary (SummHay) & 18 & .294 & .321 & .188 & +.134 \\
actions (BFCL) & 21 & .952 & .952 & .667 & +.286 \\
\midrule
\textbf{ALL} & \textbf{125} & \textbf{.778} & \textbf{.734} & \textbf{.469} & \textbf{+.265} \\
\bottomrule
\end{tabular}
\end{table}

\begin{table}[H]
\centering
\small
\caption{VitaBench interactive-transfer results ($n=100$ per condition,
Qwen3.5-122B as the tested agent and DeepSeek-V4-Flash as the user
simulator and rubric judge).}
\label{tab:appendix-vitabench-external}
\begin{tabular}{lcc}
\toprule
Condition & Score & \car{} \\
\midrule
clean (no drift) & .697 & .190 \\
drift (default close) & .720 & .190 \\
drift (no-restate close) & .594 & .190 \\
GT-oracle final intent & .764 & .310 \\
\midrule
clean $-$ no-restate drift & \multicolumn{2}{l}{$+.103$ [$+.026,+.180$]} \\
default $-$ no-restate drift & \multicolumn{2}{l}{$+.125$ [$+.050,+.202$]} \\
\bottomrule
\end{tabular}
\end{table}

The aligned no-restate condition yields a lower rubric score than clean,
while the default closing restatement masks that difference. All three
non-oracle conditions have the same \car{} (.190), so this result is
specific to partial-credit rubric evaluation in the single-model setup.

\Needspace{8\baselineskip}
\section{StateForge Diagnostics and Training Details}

\begin{table}[H]
\centering
\small
\caption{Component ablation under the unified re-run protocol ($n=125$).
Differences are paired 95\% CIs against full \stateforge{}.}
\label{tab:appendix-component-ablation}
\begin{tabular}{lccc}
\toprule
Configuration & Mean & \car{} & $\Delta$ vs full \\
\midrule
Without gate & .393 & .304 & $-.035$ [$-.101,+.030$] \\
Generic recap & .403 & .312 & $-.025$ [$-.100,+.048$] \\
Neutral recap & .441 & .344 & $+.013$ [$-.060,+.087$] \\
\stateforge{} full & .428 & .344 & --- \\
Oracle final state & .551 & .440 & $+.123$ [$+.044,+.202$] \\
\bottomrule
\end{tabular}
\end{table}

The matched recap and full-harness comparisons are not separated at this
sample size. The oracle row nevertheless shows material headroom from
exact state estimation within the history-preserving harness; it should
not be interpreted as showing that tracking dominates the gap to the
clean single-turn condition.

\begin{table}[H]
\centering
\small
\caption{Complete tracker-scale comparison on the unified re-run protocol.}
\label{tab:tracker-scale}
\begin{tabular}{lccc}
\toprule
Tracker & Mean & \car{} & $\Delta$ vs 122B \\
\midrule
2B & .156 & .112 & $-.272$ [$-.351,-.196$] \\
4B & .312 & .240 & $-.117$ [$-.199,-.036$] \\
9B & .399 & .312 & $-.029$ [$-.101,+.042$] \\
27B & .366 & .272 & $-.062$ [$-.134,+.005$] \\
35B & .395 & .304 & $-.034$ [$-.109,+.038$] \\
122B & .428 & .344 & --- \\
\bottomrule
\end{tabular}
\end{table}

For modular adaptation, the 122B base agent is frozen and paired with a
2B tracker; the reported before--after change is therefore attributable
to OPD training of the tracker within that stack. These full-episode
scores use the tracker-training evaluation protocol rather than the
\gentest{} task-score protocol.

For full-agent internalization, the 35B thinking-distilled v4 model
scores .461 on the drift set and .444 under avg@3 decoding; the base
model scores .296. The distilled model scores .718 on the clean condition
versus .745 for the base model. The v1--v4 progression, SQL-formatting
constraint, and per-family results are included in the artifact.

\Needspace{8\baselineskip}
\section{Quality-Control Rubrics}
\label{app:qc-rubrics}

We audit randomly sampled edit-phase generations rather than full traces.
The fidelity rubric checks whether an utterance realizes its scheduled
Add, Replace, Delete, or valid pre-retraction decoy without leaking a
future edit or contradicting an active requirement. The coherence rubric
checks reference resolution, active-state consistency, and dialogue flow.
Across 100 sampled utterances from the DeepSeek-V4-Flash simulator used
throughout the experiments, fidelity is 96\% and coherence is 97\%.
Residual infidelity is concentrated in redundant restatements rather
than invented requirement content.

\end{document}